\documentclass[lettersize,journal]{IEEEtran}
\usepackage{amsmath,amsfonts}
\usepackage{algorithmic}
\usepackage{algorithm}
\usepackage{array}
\usepackage[caption=false,font=normalsize,labelfont=sf,textfont=sf]{subfig}
\usepackage{textcomp}
\usepackage{stfloats}
\usepackage{url}
\usepackage{verbatim}
\usepackage{graphicx}
\usepackage{colortbl}

\usepackage{cite}
\usepackage{multirow}
\usepackage{booktabs} % For formal tables
\usepackage{url}
\usepackage{graphicx}  %Required
\usepackage{amsfonts} 
\usepackage{amsmath}
\usepackage{bm}
\usepackage{float}
\usepackage{multirow}
\usepackage{makecell}
\usepackage{xspace}
\usepackage{amssymb}
\usepackage{array}
\usepackage{xcolor}
\usepackage{mathrsfs}

\begin{document}

\title{Write Summary Step-by-Step:\\ A Pilot Study of Stepwise Summarization}

\author{Xiuying Chen$^{*}$,
        Shen Gao$^{*}$,
        Mingzhe Li,
        Qingqing Zhu,
        Xin Gao$^{\dag}$,
        Xiangliang Zhang$^{\dag}$
\thanks{Xiuying Chen and Xin Gao are with the Computer, Electrical and
Mathematical Sciences and Engineering division at KAUST, Saudi Arabia, (email: xiuying.chen@kaust.edu.sa,xin.gao@kaust.edu.sa).}%% <-this % stops a space
\thanks{Shen Gao is with the University of Electronic Science and Technology of China, China (email: shengao@sdu.edu.cn. Mingzhe Li is with Ant Group, China (email: limingzhe.lmz@antgroup.com). Qingqing Zhu is with Peking university, China (email: zhuqingqing@pku.edu.cn.).}
\thanks{Xiangliang Zhang is with University of Notre Dame, America and secondly affiliated with KAUST (email: xzhang33@nd.edu.)}%

\thanks{*These authors contributed equally to this work.}
\thanks{\dag Corresponding authors.}
}

        % <-this % stops a space

\markboth{IEEE/ACM TRANSACTIONS ON AUDIO SPEECH AND LANGUAGE PROCESSING, MANUSCRIPT}%
{Shell \MakeLowercase{\textit{et al.}}: A Sample Article Using IEEEtran.cls for IEEE Journals}


\maketitle

\begin{abstract}
Nowadays, neural text generation has made tremendous progress in abstractive summarization tasks.
  However, most of the existing summarization models take in the whole document all at once, which sometimes cannot meet the needs in practice.
  Practically, social text streams such as news events and tweets keep growing from time to time, and can only be fed to the summarization system step by step.
  Hence, in this paper, we propose the task of \textit{Stepwise Summarization}, which aims to generate a new appended summary each time a new document is proposed.
  The appended summary should not only summarizes the newly added content but is also coherent with the previous summary, to form an up-to-date complete summary.
  To tackle this challenge, we design an adversarial learning model, named \textit{Stepwise Summary Generator} (SSG).
  First, SSG selectively processes the new document under the guidance of the previous summary, obtaining polished document representation.
  Next, SSG generates the summary considering both the previous summary and the document.
  Finally, a convolutional-based discriminator is employed to determine whether the newly generated summary is coherent with the previous summary.
  \textcolor{black}{For the experiment,  we extend the traditional two-step update summarization setting to multi-step stepwise setting, and re-propose a large-scale stepwise summarization dataset based on a public story generation dataset.}
  Extensive experiments on this dataset show that SSG achieves state-of-the-art performance in terms of both automatic metrics and human evaluations. 
  Ablation studies demonstrate the effectiveness of each module in our framework.
  We also discuss the benefits and limitations of recent large language models on this task.
\end{abstract}


\section{Introduction}
\label{sec:intro}

\begin{table}
  \centering
  \caption{\textcolor{black}{Example of new document-summary pair from our dataset.
  In the previous summary, the character Stephanie claims that she can continue to work, but in the new document, she cannot focus on her work.}    
  }
  \begin{tabular}{l|l}
    \toprule
    \multicolumn{2}{p{7.5cm}}{
      \emph{\textbf{Previous Summary:}}
      Stephanie concedes that perhaps she did have a little stroke, but she is fine, still a force to be reckoned with. Nick calls for a meeting and \textit{Stephanie tells Pam that she will be there despite her protests. She needs to prove she can still do what she does. Nothing will change that.} Nick calls her a one woman dynamo.
    }\\ 
    \hline
    \multicolumn{2}{p{7.5cm}}{
      \emph{\textbf{Newly added document:}}
      ... Marcus says they were good people but ``finding mom was the greatest thing that could have happened.'' Marcus asks if he has any other kids. Justin says he tried and breaks down as he says Marcus is there before him.
      Everyone looks at numbers as Nick says they need to hire a third shift. Bridget says the Economy still has not recovered. 
      \textit{They leave the decision to Stephanie and tell her to look at page 20. She says she does not need to look at a page she knows what is best for the company and will go with her instincts.} She says ``if there is a problem with that, there's the door.'' Nick throws Jackie a look of confusion.
    }\\ 
    \hline
    \multicolumn{2}{p{7.5cm}}{
      \emph{\textbf{New Summary:}}
      However, during the meeting Stephanie can not concentrate on the figures and offers that she will do what she always has done on instinct and it's that or show her the door. Marcus wants his question answered. 
    }\\ 
    \bottomrule
  \end{tabular}
  \label{tab:intro-case}
\end{table}

First, we will answer three possible questions to illustrate the main contributions of this paper.

\subsection{What is stepwise summarization?}
Nowadays, social text streams such as news events and tweets spread throughout the Internet, and summarization is an ideal solution to provide a condensed, informative document reorganization for faster and better representation of information evolution~\cite{gidiotis2020divide}.
Most of the existing summarization approaches take in the whole document and generate the corresponding summary all at once.
However, in practice, text streams come in sequence instead of simultaneously, and it is important for readers to get the most up-to-date information each time a new event occurs.
Hence, we propose the task of \textit{stepwise summarization task}, which aims to generate a new appended summary each time a new document is proposed.
The appended summary should not only summarize the newly added content but also be coherent with the previous summary, so as to form an up-to-date complete summary.
%    When the new document arrives in a sequence, the corresponding summary needs to be updated along with, considering previous information meantime.
%    Update summarization is a well-established task, and existing update summarization approaches \cite{Delort2012DualSumAT,Mnasri2017TakingIA} are all based on human-engineered and inflexible extraction methods.
% They rely on human-engineered features and are thus not as flexible as abstractive approaches.
%    Herein, we propose the \emph{abstractive update summarization task}, which aims to concisely paraphrase the salient information in a text stream.

\begin{figure*}%
  \centering
  \includegraphics[width=1.5\columnwidth]{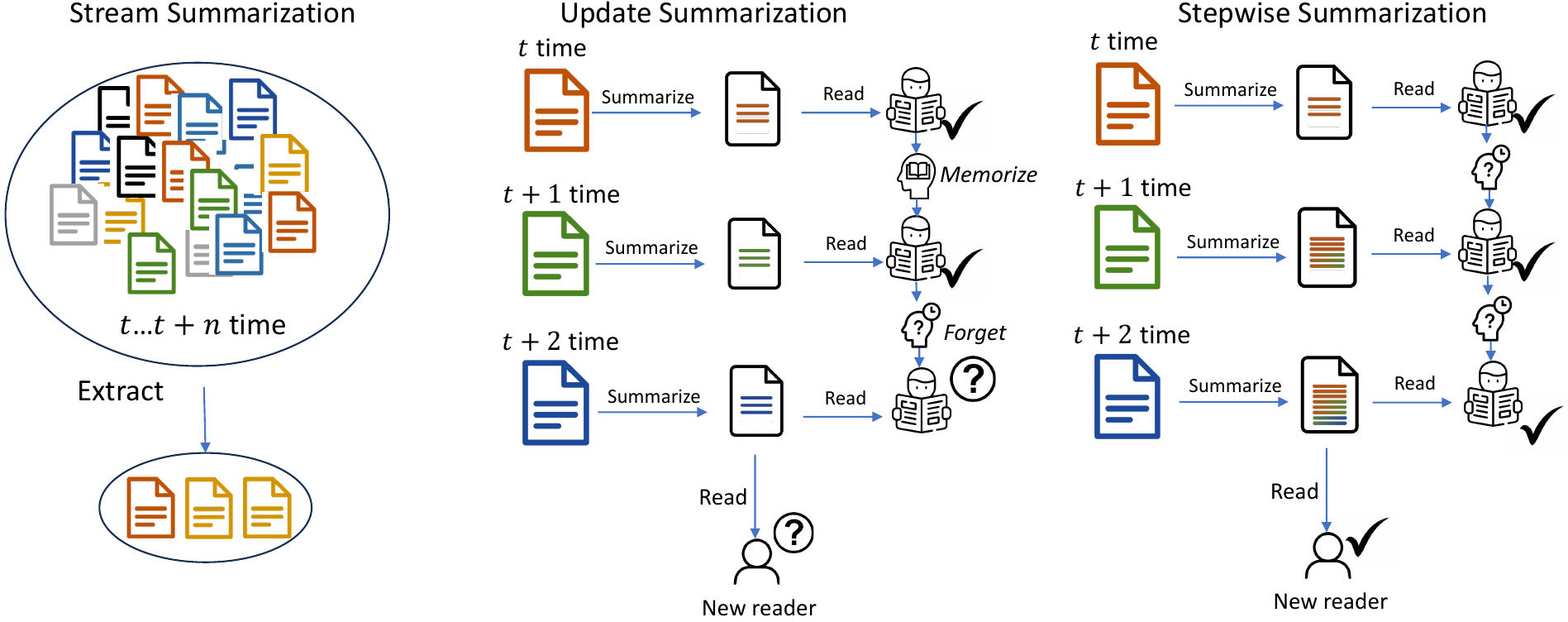}
  \caption{\textcolor{black}{Comparison between stream summarization, update summarization, and stepwise summarization.
  Stream summarization~\cite{Ge2016NewsSS}: a large number documents are taken as input, and some of the \textit{documents are selected as summary}.
Update summarization~\cite{Dang2008OverviewOT}: the task is to summarize the new document, \textit{assuming the reader already knows about previous documents}.
Stepwise summarization: This setting \textit{does not require the reader to have any prior knowledge about the text stream}. The generated summary contains all the information about the text stream, and is kept fluent all the time steps.}}
  \label{fig:overview-case2}
\end{figure*}

%    Since the input text stream keeps growing from time to time, we aim to tackle this problem by writing a new appended summary each time a new document joins the stream, i.e., writing summary step by step.
In such a scenario, the previous summary, i.e., summary of previous documents in the stream, is important in two ways.
On one hand, the newly generated summary must be coherent with the previous summary.
Considering Table~\ref{tab:intro-case} which is an episode recap collected from a fan-contributed website, the new document introduces that the character \textit{Stephanie cannot concentrate on her work}, while the previous summary describes that \textit{she says she is able to work despite her protests}.
Hence, the appended new summary summarizes the new status of Stephanie with a turning sentence. 
On the other hand, the previous summary helps capture the salient part of the new document.
% In this example, both the previous summary and the summary to be generated include information about the epidemic.
To this end, in this paper, we propose a \textit{Stepwise Summary Generator} (SSG) which incorporates both the previous summary and newly added document to generate a more coherent appended summary, so as to obtain the whole summary of a text stream.
We show an example in Figure~\ref{fig:overview-case2} to illustrate the task definition of stepwise summarization.

\subsection{Why not traditional abstractive summarization?} On one hand, if traditional summarization models only take the current newly added document as input to generate the corresponding summary, they will lose the information from previous documents that is helpful for current summarization.
The generated summaries may also be incoherent with each other.
On the other hand, if traditional models take all existing documents as input, it is easy to confuse information between events and provide inaccurate period division for readers.
It is also redundant to repeatedly generate the same summary for previous documents.
\textcolor{black}{The stepwise summarization is related to stream summarization and update summarization, where we give the comparison in Figure~\ref{sec:intro} and the related work section.}
\vspace{-2mm}

\subsection{How to conduct stepwise summarization?}
To begin with, SSG captures the main points in the document with the help of the previous summary.
Concretely, after processing the document and previous summary through self-attention, we propose a selective recurrent unit to selectively processes document representation under the guidance from the previous summary, so that document representation is further polished.
Second, a  decoder is used to generate a new summary, incorporating both the polished representations of documents and the summary.
Finally, the training of our framework is conducted in an adversarial way, where we employ a convolutional discriminator to distinguish whether the generated summary is coherent with the previous summary.
% For the experiment, we collect the first large-scale stepwise summarization dataset.
Extensive experiments  demonstrate that SSG significantly outperforms the state-of-the-art summarization baselines in ROUGE metrics and human evaluations.

Our contributions can be summarized as follows: 

$\bullet$ We propose the task of stepwise summarization, which generates an appended summary each time a new document is generated, considering previous information.

$\bullet$ To address this task, we introduce a paradigm that leverages the previous summary to enhance the current one, using an interaction and coherence discriminator.

$\bullet$ We release a large-scale stepwise summarization dataset.
Experimental results on this dataset demonstrate the effectiveness of our proposed framework.

\section{Related Work}

\subsection{Text summarization.}
Text summarization can be classified into extractive and abstractive methods. 
Extractive methods~\cite{Jadhav2018ExtractiveSW,Narayan2018RankingSF,Xu2020DiscourseAware,Dong2018BanditSum,Zhong2020Extractive} directly extract the sentences or text span from the input document as the summary.
Specifically, \cite{zhou2020joint} proposed the joint sentence scoring and selection framework which directly predicts the relative sentence importance score according to both sentence content and previously selected sentences.
However, summaries generated by extractive methods always suffer from redundancy problems and are not as flexible as abstractive methods.
Recently, with the emergence of neural network models for text generation, a vast majority of efforts have been dedicated to abstractive summarization~\cite{zhang2016abstractive,zhang2018attention,Celikyilmaz2018Deep,Wang2019BiSET,su2020two}. 
For example, \cite{Hsu2018AUM} used sentence-level attention to modulate word-level attention such that words in less attended sentences are less likely to be generated.
\cite{Gehrmann2018BottomUpAS} used a data-efficient content selector to emphasize phrases in a source document that should be part of the summary.
To tackle the out-of-vocabulary problem in abstractive summarization models, some researchers employ a copy mechanism to copy certain words from the input document to the summary~\cite{Gu2016IncorporatingCM,See2017GetTT}.

\begin{table*}
  \caption{Corpus bias towards extractive methods in the CNN/DailyMail~\cite{Hermann2015TeachingMT}, Newsroom~\cite{Grusky2018NewsroomAD}, XSum~\cite{Narayan2018DontGM}, Reddit~\cite{Kim2018AbstractiveSO} and SSD datasets. 
    We show the proportion of novel $n$-grams in gold summaries.
    We also report ROUGE scores for the extractive \texttt{Lead} and abstractive \texttt{PG} baseline.
    Results are computed on the test set. \label{tab:ngram-coverage-lead-oracle}}
  \begin{center}{
      \begin{tabular}{ l | c c c c | c c c | c c c |c} %  l l l l | l l l | l l l}
        \toprule
        \multirow{2}{*}{Datasets} & \multicolumn{4}{c|}{\% of novel n-grams in gold summary} & \multicolumn{3}{c|}{\texttt{lead}} & \multicolumn{3}{c|}{\texttt{PG}}&
        \multicolumn{1}{c}{\texttt{PG/lead}}\\
        & unigrams & bigrams & trigrams & 4-grams & R1 & R2 & RL & R1 & R2 & RL& Ratio (R-L) \\ 
        \midrule
        CNN/DailyMail & 19.50 & 56.88 & 74.41 & 82.8 & 39.6 & 17.7 & 36.2 & 36.4 & 15.7 & 33.4&0.92x \\
        Newsroom & 19.53 & 48.39 & 59.38 & 64.06 & 30.5 & 21.3 & 28.4 & 26.0 & 13.3 & 22.4&0.78x \\
        XSum & 35.76 & 83.45 & 95.50 & 98.49 & 16.3 & 1.6  & 11.9  & 29.7  & 8.8  & 22.6 &1.89x \\ 
        Reddit & 48.97 & 84.45  & 94.50 & 97.82 & 3.4 & 0.0  & 3.3  & 18.3  & 6.5 & 17.9 &5.42x \\
        SSD & 38.31 & 85.35 & 97.20 &98.86 & 26.31 & 4.79 & 23.56 & 36.83 & 9.07  & 34.37 & 1.45x\\
        \bottomrule
    \end{tabular}}
  \end{center}
\end{table*}

\subsection{Timeline summarization.}
	Timeline summarization was firstly proposed by~\cite{Allan2001TemporalSO}, and it involves extracting a single sentence from each event within a news topic.
	Since the original work, a series of works~\cite{Yan2011EvolutionaryTS,Yan2011TimelineGT,Zhao2013TimelineGW} further investigate this task.
\cite{Yan2011EvolutionaryTS} aimed to return the evolution trajectory along the timeline, emphasizing relevance, coverage, coherence, and cross-date diversity.
\cite{Chen2019LearningTA,chen2023follow} followed the style of abstractive timeline summarization.
\textcolor{black}{More recently, \cite{ghalandari2020examining} examined both the general approaches to this task and its current state of resolution, and \cite{yu2021multi} suggested enhancing timeline summarization by generating multiple summaries.}

	However, timeline summarization also belongs to the classic static summarization framework, dealing with a static corpus instead of a dynamic, continuously growing stream.

\subsection{Streaming summarization.}
Stepwise summarization is also similar to the streaming summarization task, which summarizes a stream of documents aligned on the timeline.
Initial works include \cite{shou2013sumblr}, where they aim to deal with dynamic, quickly arriving, and large-scale tweet streams. 
They proposed a prototype called Sumblr (SUMmarization By stream cLusteRing) for tweet streams.
Following their work, \cite{olariu2014efficient} improved the efficiency of this process.
Processing is done in a single pass, removing the need to save any input data and improving the running time.
\cite{ren2016time} further focused on time-aware multi-viewpoint summarization of multilingual social text streams.
% Another similar task is timeline summarization, which extracts a single sentence from each event within a news topic \cite{Allan2001TemporalSO}.
% They defined temporal summaries of news stories as extracting a single sentence from each event within a news topic, where the stories are presented one at a time and sentences from a story must be ranked before the next story can be considered.
% \cite{Yan2011EvolutionaryTS} proposed ETS, aiming to return the evolution trajectory along the timeline, consisting of an individual but correlated summary of each event.
\textcolor{black}{Both of timeline and streaming summarization tasks take stream documents all at once, while in our stepwise summarization task, the input documents come in sequence.}
%, which can be extended to multi-document summarization in the future.

\subsection{Update summarization.}
\textcolor{black}{Our task shares some common features with the update summarization task, which also aims to summarize key information from documents arriving in a sequence.}
Concretely, the setting typically includes two document sets A and B, and the task is to summarize B assuming the reader already knows about A \cite{Dang2008OverviewOT}.
The challenge in update summarization is to deal with multi-document summarization, handling redundancy between multiple documents, as well as the redundancy between A and B.
\textcolor{black}{This task differs from ours because its generated summaries do not form a coherent text sequence.}
This series of update summarization works mainly focus on TAC dataset \cite{Dang2008OverviewOT}, where the task is to extract a summary of a subsequent set of newswire articles for the same topic, under the assumption that the reader has already read the previous documents.
This dataset contains only 48 topics with 960 documents.
%    Meanwhile, this update summarization task only contains two steps, which does not simulate a real text stream that contains multi-steps.
Initial works include \cite{Boudin2008ASM}, which is a scalable sentence scoring method for query-oriented update summarization.
\cite{Delort2012DualSumAT} presented an unsupervised probabilistic approach to model novelty in a document collection and apply it to the generation of update summaries.
MCL \cite{Mnasri2017TakingIA} is the state-of-the-art extractive work on this dataset, which examines how integrating a semantic sentence similarity into an update summarization system can improve its results.

\textcolor{black}{\section{Stepwise Summarization Dataset}}
\label{sec:dataset}

\begin{figure}%
  \centering
  \includegraphics[width=0.4\columnwidth]{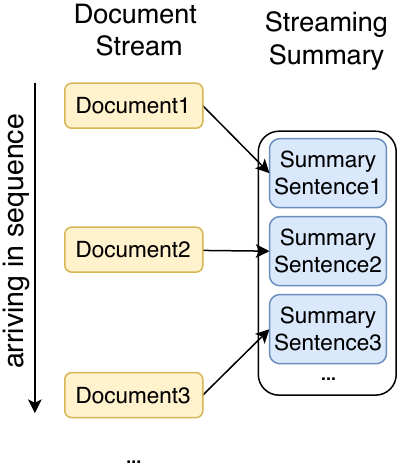}
  \caption{Illustration of a sample in the stepwise summarization dataset.}
  \label{fig:overview-case}
\end{figure}

Existing large-scale summarization datasets cannot be used for stepwise summarization, since they only contain information about a single event.
In contrast, a TV show episode recaps usually records the developments of events over time.
Thus, our dataset, named \textit{Stepwise Summarization Dataset} (SSD), is re-collected from a public story generation dataset TVMegaSite \cite{chen2022leveraging}.
As illustrated in Figure~\ref{fig:overview-case}, each case in our dataset contains a document stream and the corresponding streaming summary.
Each document stream contains several documents, and each summary stream contains the corresponding summaries.
For each doc-summ \textbf{\textit{pair}} in doc-summ \textbf{\textit{stream}}, we name all summaries and all documents before this pair as \textbf{\textit{previous} \textit{summary}} and \textbf{\textit{previous} \textit{document}}, and document in this pair as \textbf{\textit{new document}}.

\subsection{Dataset collection.}
In each case in the story generation dataset TVMegaSite, there is a detailed TV show episode recaps and a brief summary of the episode.
Since the episode recap is mostly arranged according to the sequence of events, there is a strict logic in it.
Consequently, the summary often follows the same logical order as a good-quality summary does.
Hence, an episode recap can be regarded as the streaming data source, and the summary summarizes the stream in order.
To select document-summary pairs that establish such correspondences, we first use the text segmentation technique to split the summary stream into segments.
The episode recap is separated into paragraphs in the original dataset.
Then, we calculate the mean of ROUGE-1, ROUGE-2, and ROUGE-L scores between each summary segment and document paragraphs.
We choose the section with the highest ROUGE-L score as the corresponding source document for the summary segment.
For each source paragraph, the summary segment that has the highest ROUGE score with it is selected as its corresponding summary.
A document stream with less than two paragraphs will also be filtered.
Overall, SSD consists of 29,376 doc-summ pair samples, 5,208 validation samples, and 5,718 test samples.
A document consists of 610 words, and a summary contains 87 words on average.
Document streams containing 2,3,4, and 5 documents make up 35\%, 22\%, 18\%, and 5\% of the total respectively.

Our dataset is notably the first large-scale stepwise summarization dataset and will be released for further research. 
Detailed statistics for SSD and a comparison with other popular text summarization datasets are listed in Table~\ref{tab:ngram-coverage-lead-oracle}.
\texttt{Lead} selects the first sentence for Reddit dataset, and the first three sentences for others.
Next, we demonstrate the quality of the dataset and then discuss some abstractive characteristics of SSD compared to existing summarization datasets.

\begin{figure*}%
  \centering
  \includegraphics[width=1.6\columnwidth]{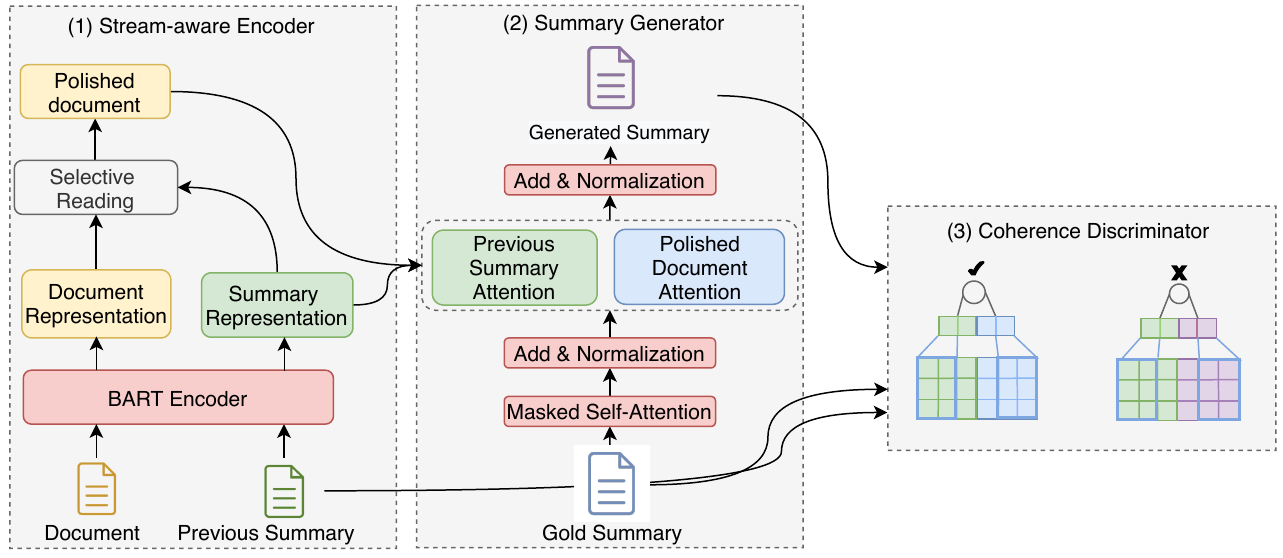}
  \caption{\textcolor{black}{Overview of SSG. We divide our model into three parts: (1) \textit{Stream-aware Encoder} (left) selectively processes document under guidance from the previous summary. (2) \textit{Summary generator} (middle) generates a summary taking both factors above into consideration. (3) \textit{Coherence discriminator} (right) determines whether the generated summary is coherent with the previous one.}}
  \label{fig:overview}
\end{figure*}

\subsection{Stepwise summarization characteristics.}
\textbf{Human evaluation. }
We assess the quality of our stepwise summarization data from two aspects: whether the source summary corresponds to the document, and whether the previous summary relates to the new document.
The first aspect reflects the quality of the  \textit{summarization attribute} and the second reflects the \textit{stream attribute}.    
Summarization attribute here means whether the summary can correctly summarize important information from document, and stream attribute means whether the previous information relates to current document.
We randomly sample 100 cases from SSD and perform Amazon’s Mechanical Turk tests, where three annotators are asked to assess the aforementioned aspects for each selection, i.e., with ``yes'' and ``no''.
The result shows that ``yes'' accounts for 94.3\% and 95.0\% of the answers for the summarization attribute and stream attribute, while ``no'' only takes up 5.7\%, and 5.0\%, respectively.
This demonstrates that the dataset does indeed have good summarization and stream attributes.
The kappa statistic is 0.57, which demonstrates moderate agreement between turkers.

\textbf{Automatic evaluation. }
We also propose an automatic method to evaluate the dataset, since the original TVMegaSite dataset provides all the mentioned character names.
Ideally, if a character exists in a summary, he/she should also be included in the source paragraph, and vice versa.
Hence, we calculate the percentage of cases when a character name exists in the paired document and summary (good case), and the cases when it only exists in the summary (bad case).
The result shows that 89\% of cases are good cases.
This again demonstrates the good quality of the dataset.

\subsection{Abstractive characteristics.}
SSD has various abstractive characteristics, including strong abstractness and weak lead bias, as shown in Table~\ref{tab:ngram-coverage-lead-oracle}.
About 38.31\% unigrams and 96.68\% 4-grams in the gold summary are novel n-grams that do not exist in the input document, indicating strong abstractness in the dataset.
This leads to a necessity for abstractive summarization methods since these n-grams cannot be generated by extractive models.
What is more, SSD has a weak lead bias.
The \texttt{Lead} baseline is to select the first three sentences in the document as a summary, and the \texttt{PG} baseline corresponds to the strong abstractive summarization baseline \cite{See2017GetTT}.
The higher the \texttt{PG/Lead} score, the weaker the lead bias of the dataset.
News datasets such as CNN/DailyMail and Newsroom always have a strong lead bias, with a \texttt{PG/Lead} ratio of less than one.
Datasets such as XSum and Reddit overcome this problem.
Our dataset also has a weak lead bias, where the \texttt{PG/Lead} ratio is larger than one.

\section{Problem Formulation}
\label{sec:formulation}

Before presenting our approach for the stepwise summarization, we first introduce our notations and key concepts. 

To begin with, for a document $X^d=\{x^d_1, x^d_2, \dots, x^d_{T^d}\}$, we assume there is a previous document-summary pair, where the previous summary is $X^s=\{x^s_1, x^s_2, \dots, x^s_{T^s}\}$. $x^d_i$ denotes the $i$-th word in document $X^d$, and $x^s_{i}$ denotes the $i$-th word in previous summary $X^s$.
Given the document $X^d$ and previous summary $X^s$, the summary generator generates a summary $\hat{Y} = \{\hat{y}_1, \hat{y}_2, \dots, \hat{y}_{T^{\hat{Y}}}\}$.
Finally, we use the difference between the generated summary $\hat{Y}$ and the ground truth summary $Y = \{y_1, y_2, \dots, y_{T^{Y}}\}$ as the training signal to optimize the model parameters.

\section{The Proposed SSG Model}

\subsection{Overview}
In this section, we propose the \emph{Stepwise Summary Generator}, abbreviated as SSG. 
An overview of SSG is shown in Figure~\ref{fig:overview}, which can be split into three main parts:

$\bullet$ \textit{Stream-aware encoder} incorporates the previous summary to calculate the selective reading gate in the gated recurrent unit, further polishing the document representation.

$\bullet$ \textit{Summary generator} generates a summary taking both the polished document and previous summary into consideration.

$\bullet$ \textit{Coherence discriminator} uses a convolutional neural network (CNN) to identify whether the generated summary is coherent with the previous summary.

\subsection{Stream-aware Encoder}

We first use an embedding matrix $e$ to map a one-hot representation of each word in $X^d$ and ${X^s}$ into a high-dimensional vector space.
We then employ the Transformer encoder \cite{Vaswani2017AttentionIA} initialized by BART~\cite{lewis2020bart} to model the temporal interactions between words, and obtain $h^d_t$ and $h^s_t$, denoting the hidden state of the $t$-th word in $X^d$, $X^s$ respectively. 
% Following~\cite{See2017GetTT}, we choose Long Short-Term Memory (LSTM) as the cell for Bi-RNN.
% LSTM can be replaced with similar algorithms, such as Gated Recurrent Unit (GRU)~\cite{chung2014empirical}, but we leave this as future work.

To further model the interaction between a document and previous summary, we propose to utilize the previous summary to selectively read and process the document, in order to emphasize the words that are related to the previous summary.
This is achieved by an RNN made up of Selective Reading Unit (SRU).
% , as shown in Figure~\ref{fig:encoder}.
Our inspiration comes from \cite{Chen2018IterativeDR}, which iteratively polished the document representation using document words.
However, in SSG, we use the previous summary representation to polish the word representations in the document.

We begin by introducing the original GRU cell, which consists of an update gate vector ${u^g_t}$ and a reset gate vector ${r^g_t}$.
In our case, the output is the polished representation of ${h^d_t}$, which is also the hidden state in the SRU cell, and we denoted as $h^g_t$.
For each time step $t$ with the document representation $h^d_t$ and the previous hidden state ${h^g_{t-1}}$, the updated hidden state ${h^g_{t}}=\text{GRU}({h^d_t},{h^g_{t-1}})$ is computed by:
\begin{align}
{u^g_{t}} &=\sigma({W_1}{h^d_t}+{U_1}{h^g_{t-1}}+{b_1}) , \label{gated} \\
{r^g_{t}} &=\sigma({W_2}{h^d_t}+{U_2}{h^g_{t-1}}+{b_2}) , \\
{\tilde{h}^g_{t}}&=\text{tanh}({W_3 h^d_t}+{r^g_{t}}\circ {U_3h^g_{t-1}}+{b_3}), \\
{h^g_{t}}&={u^g_{t}}\circ {\tilde{h^g_{t}}}+(1-{u^g_{t}})\circ {h^g_{t-1} }, \label{hi}
\end{align}
where $W$ and $b$ are trainable parameters, $\sigma$ is the sigmoid activation function, and $\circ$ is element-wise multiplication.
Here, due to the way $u_t^g$ is calculated, it is sensitive to the position and order of the current input document, but loses information from the previous summary.
In SRU, we propose to take the previous summary into consideration when calculating $u_t^g$:
\begin{align}
{f_{t}} & =[h^d_t\circ {q}; {h^d_t}; {q}], \\
{F_{t}}& ={W_4}\text{tanh}({W_5}{f_{t}}+{b_5})+{b_4}, \\
{u'_{t}}& =\frac{\text{exp}({F_{t}})}{\sum^{T^d}_{j=1}\text{exp}({F_{j}})},
\end{align}
where $[;]$ denotes concatenation, ${h^d_t}$ is the $t$-th word representation, and ${q}$ is the average of the previous summary representations $h^s_\cdot$. Equation \ref{hi} now becomes:
\begin{align}
{h^g_{t}}&={u'_{t}}\circ {\tilde{h}^g_{t}}+(1-{u'_{t}})\circ {h^g_{t-1} }.
\end{align}
In this way, SSG automatically decides the extent to which the information of each word should be updated based on its relationship with the previous summary.

\textcolor{black}{\subsection{Summary Generator}}

The summary generator aims to generate a summary based on the input document and the previous summary.
We also employ the classic Transformer architecture as the decoder.
Generally, each layer in the decoder consists of three parts: a multi-head self-attention mechanism, a multi-head cross-attention mechanism, and a fully connected feed-forward network.

For each layer, at the $t$-th decoding step, we first apply the self-attention on the masked summary embeddings, denoted as $g_t$.
The self-attention sub-layer with a masking mechanism is used to encode the decoded information. 
The masking mechanism ensures that the prediction of the position $t$ depends only on the known output of the position before $t$.
\begin{align}
    \tilde{g}_t=\text {LN }\left(g_t+\text{MSAttn}\left(g_{<t}, g_{<t}\right)\right),
\end{align}
where LN is the layer normalization operation; MSAttn is the masked multi-head attention operation.
Based on $\tilde{g}_t$ we compute the cross-attention scores over polished document:
\begin{align}
z_{a,t}=\text{ReLU}([\tilde{g}_t W_a(h^g W_b)^T]),
\end{align}
where $W_a, W_b \in \mathbb{R}^{d \times d}$, $d$ is the hidden dimension, $h^g$ is the concatenation of all $h^g_i$ for $i \in [1, T_d]$, and $z_{a,t} \in \mathbb{R}^{T^d}$.
Similarly, we also calculate the cross-attention scores over the previous summary:
\begin{align}
z_{b,t}=\text{ReLU}([\tilde{g}_t W_c(h^s W_d)^T]),
\end{align}
where $z_{b,t} \in \mathbb{R}^{T^s}$.

{The attention weights $z_{a,t}$ and $z_{b,t}$ are then used to obtain the context vectors $c_{a,t}$ and $c_{b,t}$, respectively.}
Take the document context vector as an example:
\begin{align}
    c_{a,t}= z_{a,t} h^g.
\end{align}

These context vectors, treated as salient contents summarized from various sources, are concatenated with the decoder hidden state $\tilde{g}_t$ to produce the distribution over the target vocabulary:
\begin{align}
    P_t^{\text {vocab }}=\text{Softmax}\left(W_{o}\left[\tilde{g}_t ; c_{a,t};c_{b,t}\right]\right).
\end{align}
All the learnable parameters are updated by optimizing the negative log likelihood objective function of predicting the target words:
\begin{align}
    \mathcal{L}_{s}=-\textstyle \sum_{t=1}^{T^y} \log P_t^{\text {vocab}}\left(y_{t}\right).
\end{align}

\subsection{Coherence Discriminator}

Previous work~\cite{Holtzman2018LearningTW} has found that using a cross-entropy loss alone is not enough for generating coherent text.
Hence, we employ a discriminator to provide additional training signals for the summary generator.
The goal of this discriminator is to determine whether the new summary is coherent with the previous one.
The generated summary is treated as the negative sample and the ground truth answer as the positive one.

Similar to before, we use the same embedding matrix and another attention mechanism to process $\hat{y}$ as $h^y_t$.
Since the ground truth is encoded by a transformer that is different from the decoder transformer, we use a linear projection to transform the high-dimensional space of $h^y_t$ to the same space as the decoder hidden state $g_t$:
\begin{align}
d^y_t &= W_z h^y_t +b_z.
\end{align}

We model the interaction between the previous summary and generated one by concatenating them as $\hat{d}^y_{t}=[h^s_t;d^y_{t}]$, and the previous summary and ground truth summary as $\hat{d}^g_{t}=[h^s_t;g_{t}]$.
Then, a convolutional layer convolves $\hat{d}^*_{t}$ with multiple convolutional kernels of different widths, obtaining a sequence of new features $n^*_{t}$:
\begin{align}
n^*_{t} &= \text{relu}(\hat{d}^*_{t} \odot W_c+b_c),
\end{align}
where $\odot$ denotes the convolution operation.
For each convolutional filter, the max-pooling layer takes the maximal value among the generated convolutional features $n^y_{}$ and $n^g_{}$, respectively, resulting in a fixed-size vector $N^y$ and $N^g$.
Finally, we obtain the classification result $D(\hat{d}^*)$:
\begin{align}
D(\hat{d}^*) &= \sigma(W_h\text{relu}(N^*)+b_h).
\end{align}

The training objective of the discriminator is to maximize the log-likelihood for classification, while the generator aims to maximize the probability of assigning the correct label to the generated summary:
\begin{align}
\mathcal{L}_d &=  -\left( \log(D(\hat{d}^y)) + \log(1-D(\hat{d}^g)) \right),\\
\mathcal{L}_g &=  \log(1-D(\hat{d}^g)).
\end{align}

\textcolor{black}{Overall, the parameters in the convolutional discriminator are optimized by the loss function $\mathcal{L}_d$, while other parameters are optimized by $\mathcal{L}_g$.}

\section{Experimental Setup}
\label{sec:experiment}

\subsection{Evaluation Metrics}
For evaluation, we adopt ROUGE scores~\cite{Lin2004ROUGEAP} which are widely applied to assess summarization quality~\cite{Hsu2018AUM,Narayan2018RankingSF}. 
The ROUGE metrics compare a generated summary with a reference summary by computing overlapping lexical units.
We report unigram and bigram overlap (ROUGE-1 and ROUGE-2) as a means of assessing the informativeness and the longest common subsequence (ROUGE-L) as a means of assessing fluency.
We then use BERTScore \cite{zhang2019bertscore} to calculate a similarity score between the summaries based on their BERT embeddings.
Since only using automatic evaluation metrics can be misleading~\cite{Stent2005EvaluatingEM}, we also conduct the human evaluation. 

\subsection{Comparison Methods}

To evaluate the performance of our proposed model, we compare it with the following baselines:

\texttt{Lead3} selects the first three sentences of a document as the summary.
\texttt{TextRank}: \cite{Mihalcea2004TextRankBO} proposes to build a graph, then adds each sentence as a vertex and uses links to represent semantic similarity. 
 \texttt{MCL}: \cite{Mnasri2017TakingIA} integrates semantic sentence similarity into an update summarization system.
  The proposed model is based on ILP \cite{Gillick2009ASG}, and is easily adapted to a single-document summarization setting.
 \texttt{Transformer} is based on attention mechanism proposed in~\cite{Vaswani2017AttentionIA}.
 \texttt{PG} combines the sequence-to-sequence framework with the copy mechanism from~\cite{See2017GetTT}.
  \texttt{SAGCopy}: a model proposed by \cite{Xu2020SelfAttentionGC}, which augments Transformer with self-attention guided copy mechanism.
   \texttt{BART}: a denoising autoencoder for pretraining sequence-to-sequence models \cite{lewis2020bart}.
  \textcolor{black}{  \texttt{PEGASUS}: a model pre-trained with extracted gap-sentences for abstractive summarization \cite{zhang2020pegasus}.
  Our model is implemented based on \texttt{PEGASUS} and \texttt{BART} due to their better performance.
   We also propose two simple but intuitive models based on \texttt{BART} and \texttt{PEGASUS} model, i.e., \texttt{Concat-prev-summ (CPS)}, which concatenates a new document with its previous summary, and \texttt{Concat-prev-doc (CPD)} which concatenates new document with the previous document.
   For human evaluation, we also add the performance of the latest large language model, GPT-3.5.}

\subsection{Implementation Details}

We implement our experiments in PyTorch on an NVIDIA A100 GPU.
The model is finetuned based on BART-base.
The encoding step is 1000 for a document and 300 for a previous summary.
We chose 1000 as our truncation size as we did not find significant improvement when increasing the input length from 1000 to 1500 tokens.
The minimum decoding step is 100, and the maximum step is 300.
The batch size is set to 32.
We use Adam optimizer~\cite{kingma2014adam} as our optimizing algorithm.
	We also apply gradient clipping~\cite{Pascanu2013OnTD} with a range of $[-2,2]$ during training. 
For the test, we employ beam search with a beam size of 4 to generate more fluent summaries.

\textcolor{black}{\section{Experimental Results}}

\textcolor{black}{\subsection{Overall Performance}}

\begin{table}[t]
  \centering
  \caption{ROUGE score comparisons between baselines in doc-summ pair level.
    All our ROUGE scores have a 95\% confidence interval of at most $\pm$0.23 as reported by the official ROUGE script.}
  \begin{tabular}{@{}l cccc @{}}
    \toprule
    & RG-1 & RG-2 & RG-L & BS \\
    \midrule
    \multicolumn{4}{@{}l}{\emph{Sentence extraction methods}}\\
    Lead3 & 26.31 & 4.79 & 23.56 & 86.01\\
    TextRank & 32.71 & 6.49 & 29.06 & 85.90 \\
    MCL & 33.57 & 6.73& 29.96 & 85.88\\
    \midrule
    \multicolumn{4}{@{}l}{\emph{Other previous information utilization methods}}\\
    CPD &33.67& 7.89&30.60 & 86.57 \\
    CPS & 33.97 & 7.94 & 30.75& 86.65  \\
    \midrule
    \multicolumn{4}{@{}l}{\emph{Text generation methods}}\\
    Transformer & 26.24& 5.56& 24.79& 86.04 \\
    PG &  27.18 & 5.82 & 25.01 & 86.12\\
    SAGCopy & 28.72 &6.44 &25.82  & 86.39\\
    BART &  33.30 &7.55 &  30.18& 86.42\\
 \textcolor{black}{PEGASUS} & \textcolor{black}{32.35}& \textcolor{black}{6.91} & \textcolor{black}{29.78}& \textcolor{black}{86.10}\\
 \textcolor{black}{GPT3.5} & \textcolor{black}{33.67}& \textcolor{black}{6.39} & \textcolor{black}{31.03}& \textcolor{black}{86.23}\\
    \textcolor{black}{SSG (PEGASUS)} & \textcolor{black}{34.23} & \textcolor{black}{8.08} & \textcolor{black}{31.24} & \textcolor{black}{86.54} \\
    SSG (BART) & \textbf{34.92} & \textbf{8.65} & \textbf{31.68} & \textbf{86.96} \\
    
    \bottomrule
  \end{tabular}
  \label{tab:comp_rouge_baselines}
\end{table}

We first examine \textit{whether our stepwise summarization model outperforms traditional summarization models}.
In this setting, performances are evaluated in doc-summ pair level (in contrast to doc-summ stream level).
Table~\ref{tab:comp_rouge_baselines} lists the performances of all comparison methods in terms of ROUGE scores.
Firstly, extractive methods are competitive with abstractive baselines in terms of ROUGE score but underperform in BERTScore.
This indicates that the extracted summaries have lower semantic similarity compared with abstractive summaries.
Secondly, abstractive models equipped with pretrained language models outperform traditional summarization models.
\textcolor{black}{GPT-3.5 is competitive in a text-pair setting.}
Finally, \texttt{SSG} achieves consistently better performance, with 4.86\% and 4.97\% improvements over \texttt{BART} in ROUGE-1 and ROUGE-L.

We next examine \textit{how effective is previous information in stepwise summarization}.
Performances of these \texttt{CPS} and \texttt{CPD} are relatively higher than \texttt{BART} but worse than \texttt{SSG}, demonstrating that previous information is indeed helpful for summary generation even by simple concatenation methods, but this additional information is not the only reason why \texttt{SSG} can achieve good performance.
\texttt{CPS} gives better results than \texttt{CPD}, which proves the superiority of using the previous summary over using previous documents.
%	However, neither performs as well as any of the ablation models of SSG in Table~\ref{tab:comp_rouge_ablation}, demonstrating that our proposed model successfully takes advantage of previous information.

\textcolor{black}{\subsection{Ablation Study}}

\begin{table}[t]
  \centering
  \caption{ROUGE scores of different ablation models.}
  \begin{tabular}{@{}lcc cc@{}}
    \toprule
    & ROUGE-1 & ROUGE-2 & ROUGE-L & BS\\
    \midrule
    SSG w/o SR & 34.01 &8.20 &  31.04 & 86.57 \\
    SSG w/o GAN &34.38  &8.41  &31.28 &86.86 \\
    \midrule
    SSG & \textbf{34.92} & \textbf{8.65} & \textbf{31.68} & \textbf{86.96} \\
    \bottomrule
  \end{tabular}
  \label{tab:comp_rouge_ablation}
\end{table}

Next, we conduct ablation tests to assess the importance of the selective reading module (\texttt{w/o SR}), as well as the discriminator (\texttt{w/o GAN}), and the ROUGE score results are shown in Table~\ref{tab:comp_rouge_ablation}.
The discriminator provides the scalar training signal $\mathcal{L}_d$ for generator training.
Consequently, there is an increase in performance of 1.27\% from SSG w/o GAN to SSG in terms of ROUGE-L, which demonstrates the effectiveness of the discriminator.
As for the selective reading module, it improves performance by 2.06\% compared with SSG w/o SR, in terms of ROUGE-L, indicating that a deeper interaction between the previous summary and document further helps summarization.

\textcolor{black}{\subsection{Performance on Text Stream}}
Note that previous experiments are based on generating a single summary, whereas previous summaries are ground truth summaries.
To fully simulate a real text stream scene, we also conduct experiments where the target is to generate the whole summary stream, as shown in Figure~\ref{fig:overview-case}.
The difference lies in that SSG now takes a previously generated summary as the previous summary rather than the ground truth summary.
We select several strong abstractive models in Table~\ref{tab:comp_rouge_baselines} as baselines.
As stated in the introduction, there are two ways of generating a summary stream for traditional abstractive models.
The first involves taking all documents and generating all summaries together, while the second generates a summary step-by-step.
We denote the variations following the first fashion as \texttt{*-Together}, and the second as \texttt{*-Split}.

The comparison results are listed in Table~\ref{tab:together}.
We can see that overall, the performance of each model under the stepwise summarization setting is similar to its performance under the original setting.
\texttt{*-Split} methods are better than \texttt{*-Together}, which may be because \texttt{*-Together} tend to mix the information between documents, while \texttt{*-Split} models can accurately utilize current document information, even when they do not have access to knowledge from previous documents.
However, none of these models perform as well as SSG, since it can utilize current document information, while also capturing the relationship between the current and previous summary.
\textcolor{black}{GPT-3.5 falls behind in a text-stream setting, likely due to difficulty in handling sequential information.}
\textcolor{black}{Note that in this setting, the ROUGE score is calculated in terms of text stream level, thus the results are not directly comparable to results in Table~\ref{tab:comp_rouge_baselines}.}

\subsection{Human Evaluation and Case Study}
We also assessed system performance on the streaming text by human judgments on 30 randomly selected instances following \cite{Liu2019HierarchicalTF}.
Our first evaluation study quantified the degree to which summarization models retain key information from the documents following a question-answering (QA) paradigm.
The evaluation size is 1.5 times larger than the original setting~\cite{Liu2019HierarchicalTF}.
We created a set of questions based on the multiple gold ground summaries and examined whether participants were able to answer these questions by reading system summaries alone.
We created 61 questions in total. 
Examples of questions and their answers are given in Table~\ref{tab:case}.
We adopted the same scoring mechanism used in \cite{Clarke2010DiscourseCF}, i.e., correct answers are marked with 1, and 0 otherwise.
Our second evaluation assessed the overall quality of the summaries by asking participants to rate them taking into account the following criteria: \textit{Informativeness}, \textit{Consistency}, and \textit{Succinctness}. 
The rating score of each system ranges from 1 (worst) to 5 (best).

Both evaluations were conducted by three Ph.D. students. 
Participants evaluated summaries produced by \texttt{SAGCopy-Split}, \texttt{BART-Split}, \texttt{GPT-3.5}, and our \texttt{SSG}.
All evaluated systems were variants that achieved high performance in automatic evaluations. 
As shown in Table~\ref{tab:comp_human_baslines}, on both evaluations, participants overwhelmingly prefer our model compared with baselines of similar scales. 
All pairwise comparisons among systems are statistically significant using the paired student t-test for significance at $\alpha$ = 0.01.

\begin{table}[t]
  \centering
\textcolor{black}{  \caption{ROUGE score comparisons on text stream. ROUGE score comparisons between baselines.
    All scores have a 95\% confidence interval of at most $\pm$0.28.
    }}
  \begin{tabular}{@{}lcc c c@{}}
    \toprule
    & RG-1 & RG-2 & RG-L & BS \\
    \midrule
    \multicolumn{4}{@{}l}{\emph{*-Together methods}}\\
    Transformer-Together& 36.17 & 8.76 & 33.45& 82.29 \\
    PG-Together & 36.36 & 8.79 & 33.95 & 82.41 \\
    SAGCopy-Together & 37.89 &9.50  &35.54& 82.74\\
    \textcolor{black}{PEGASUS-Together} &\textcolor{black}{45.05}  &\textcolor{black}{10.76} &\textcolor{black}{42.74} & \textcolor{black}{84.22}\\
    BART-Together & 45.36 &11.15 & 43.14 & 85.34\\
    \midrule
    \multicolumn{4}{@{}l}{\emph{*-Split methods}}\\
    Transformer-Split& 36.30 & 8.93 & 33.74& 82.31 \\
    PG-Split & 36.83 & 9.07&   34.37 &83.50 \\
    SAGCopy-Split & 38.44 & 9.24  &34.99& 83.81\\
     \textcolor{black}{PEGASUS-Split } & \textcolor{black}{45.19} & \textcolor{black}{10.79}&\textcolor{black}{43.16}& \textcolor{black}{84.77}\\
      \textcolor{black}{GPT-3.5} & \textcolor{black}{34.20} & \textcolor{black}{7.11}&\textcolor{black}{30.85}& \textcolor{black}{80.06}\\
    BART-Split & 45.94 &11.28  & 43.96 & 84.69\\
    SSG & \textbf{47.00} & \textbf{12.46} & \textbf{44.77} & \textbf{85.86}\\
    \bottomrule
  \end{tabular}
  \label{tab:together}
\end{table}

We give two examples of system outputs in Table~\ref{tab:case}.
In the first case, we can see that our proposed model \texttt{SSG} can generate a coherent, informative, and accurate summary.
Both baselines make some mistakes or miss important information.
In the second case, all baselines and our model make some mistakes. 
\texttt{BART-Split} confuses the father and son, and our \texttt{SSG} introduces ``the son goes to the dad'' while the fact is ``the dad goes to the son''.
\textcolor{black}{It's evident that GPT-3.5 generally performs comparably to our model, with the exception being its misinterpretation of ``leave Nikolas a widower'' as ``causing Nikolas's death''.}

\begin{table}[t]
  \centering
  \caption{\textcolor{black}{System scores based on questions answered by human annotators and summary quality rating.}}
  \begin{tabular}{@{}lcc@{}}
    \toprule
    & QA(\%) & Rating \\
    \midrule
    SAGCopy-Split &34.42 &3.26 \\
    BART-Split & 40.98 & 3.56\\
    SSG & \textbf{45.90 }&\textbf{3.81} \\
    GPT-3.5 & 48.39 &4.03 \\
    \bottomrule
  \end{tabular}
  \label{tab:comp_human_baslines}
\end{table}

\begin{table}
  \centering
  \caption{\textcolor{black}{Gold human authored summaries and automatic summaries produced by several baselines and our \texttt{SSG}.}
  }
  \resizebox{0.5\textwidth}{!}{
    \begin{tabular}{l|l}
      \toprule
      \multicolumn{2}{p{7.5cm}}{
        \emph{\textbf{Ground Truth Summary:}}
        Back at home, the not-so-happy newlyweds get into an argument which prompts Lydia to pick up the phone and call a lawyer so she can have the marriage annulled. However, Stefan catches her and hangs up the phone, telling her that the only way she'll get out of the marriage is by leaving Nikolas a widower. 
      }\\ 
      \hline
      \multicolumn{2}{p{7.5cm}}{
        \emph{\textbf{QA:}} Question: Who stops Lydia from calling a lawyer? Answer: Stefan.
      }\\
      \hline
      \multicolumn{2}{p{7.5cm}}{
        \emph{\textbf{SAGCopy-Split:}} \underline{Emily} and Stefan get into an argument which prompts Lydia to call a lawyer so she can have the marriage annulled. Stefan catches her and hangs up the phone, telling her that the only way she'll get out of the marriage is by leaving Nikolas a widower.  
      }\\
      \hline
      \multicolumn{2}{p{7.5cm}}{
        \emph{\textbf{BART-Split:}} Lydia and Stefan get into an argument which prompts Lydia to pick up the phone and call a lawyer so she can have the marriage annulled. However, Stefan catches her and hangs up.
      }\\
      \hline
      \multicolumn{2}{p{7.5cm}}{
        \emph{\textbf{GPT-3.5:}} \textcolor{black}{Lydia and Nikolas have a heated argument, leading Lydia to seek an annulment through a lawyer. Stefan intervenes, insisting that the only way  \underline{she can escape the marriage is by causing Nikolas's death.}}
      }\\
      \hline
      \multicolumn{2}{p{7.5cm}}{
        \emph{\textbf{SSG:}
        }
        Lydia and Stefan get into an argument which prompts Lydia to call a lawyer so she can have the marriage annulled. However, Stefan catches her and hangs up the phone, telling her that the only way she'll get out of the marriage is to leave Nikolas a widower. 
      }\\
      \toprule
      \multicolumn{2}{p{7.5cm}}{
        \emph{\textbf{Ground Truth Summary:}}
        \underline{Sonny goes to intervene} in what could have been an escalated situation involving Morgan's confrontation with a guy at the coffee shop. When it appears his son handled the situation and nothing got out of hand, Sonny commends and praises him, tells Morgan how proud he is of him and assures his son all will be well as long as he takes the correct meds that have proven successful.
        % Meanwhile, Ava goes to Sonny and Carly's home to pick up Avery and take her home.
      }\\ 
      \hline
      \multicolumn{2}{p{7.5cm}}{
        \emph{\textbf{QA:}} Question: Who is the father? Answer: Sonny.
      }\\
      \hline
      \multicolumn{2}{p{7.5cm}}{
        \emph{\textbf{SAGCopy-Split:}} Sonny talks to \underline{Morgan who encourages his son} to know that he did the right thing, reminding him he protected Kiki without losing control. He tells Morgan he needs to keep doing what he is doing and continue taking his meds and he will be all right. 
      }\\
      \hline
      \multicolumn{2}{p{7.5cm}}{
        \emph{\textbf{BART-Split:}}At the coffee shop, \underline{Sonny talks to his dad} who encourages his son to know that he did the right thing by protecting Kiki without losing control. He assures his son that the behavior of the guy and the situation was enough to reasonably provoke and piss anybody off. Sonny assures Morgan he merely needs to keep doing what he is doing and will be all right. 
      }\\
      \hline
      \multicolumn{2}{p{7.5cm}}{
        \emph{\textbf{GPT-3.5:}} \textcolor{black}{In the article, Sonny advises Morgan to believe he did the right thing by protecting Kiki from a provoking individual. Sonny reassures Morgan that as long as he continues taking his medication and stays on the right path, he will be fine.}
      }\\
      \hline
      \multicolumn{2}{p{7.5cm}}{
        \emph{\textbf{SSG:}
        }
       \underline{Morgan goes to talk to his dad} who encourages his son to know that he did the right thing by protecting Kiki from the guy who provoked him. They both affirm that Morgan is successfully fighting his illness and getting his life back on track. 
      }\\
      \bottomrule
  \end{tabular}
  }
  \label{tab:case}
  \vspace{-5mm}
\end{table}

\textcolor{black}{\subsection{Influence of Previous Information}}

\begin{figure}
  \centering
  \includegraphics[scale=0.26]{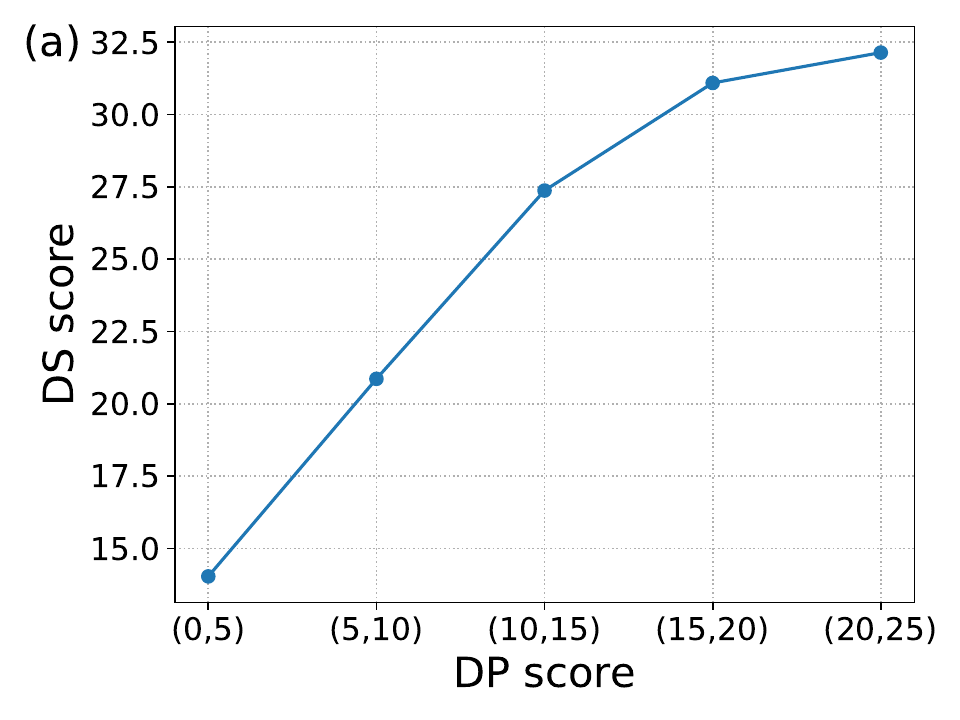}
  \includegraphics[scale=0.26]{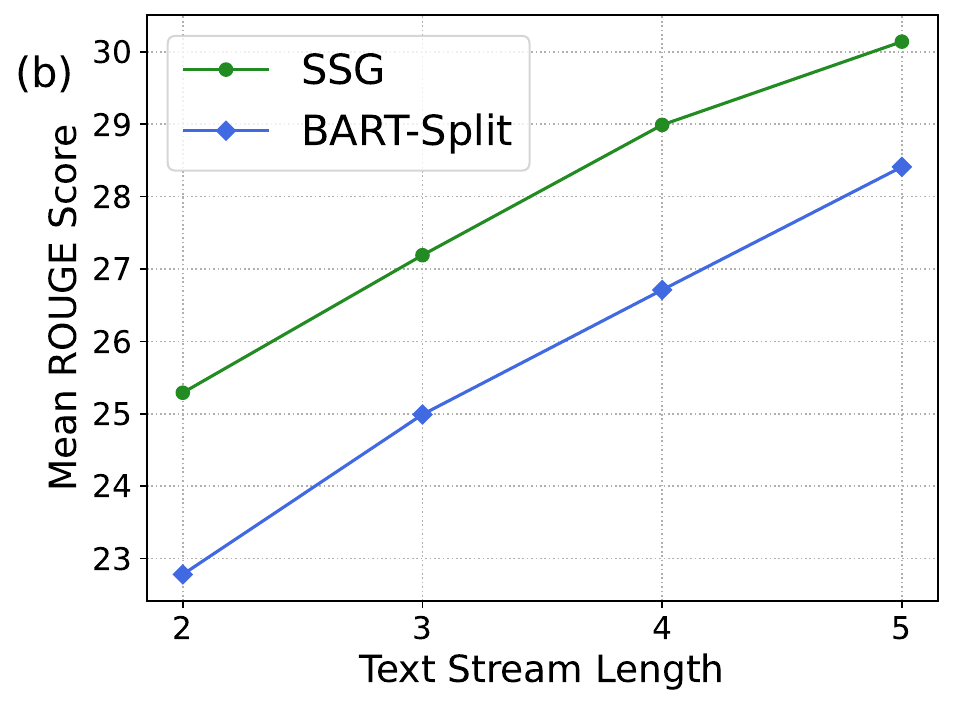}
  \caption{\textcolor{black}{(a) Influence of previous information (DP score) on summarization performance (DS score). (b) Error accumulation analysis when the text stream gets longer.}}
  \label{figures}
\end{figure}

Since our stepwise summarization takes the previous information as input, we conduct an experiment to see the influence of the relatedness between the previous and current information.
Concretely, we calculate the ROUGE score between the document and generated summary (DS score), and the score between the document with previous summary (DP score).
The result is shown in Figure~\ref{figures}(a).
It can be seen that the two scores have a positive correlation with each other, which means that a closer relationship between the previous and current information can help improve summarization performance.
However, the DS scores are substantially higher than DP scores, which means that the good performance of our model does not solely rely on the previous information, but also on the modeling of the two information sources.

\textcolor{black}{\subsection{Error Accumulation Analysis}}

As the text stream gets longer, the errors in the summarization process get accumulated.
Hence, we are interested to see whether the benefits of stepwise summarization will be diminished as the stream grows.
We conduct experiments to investigate the ROUGE performance on text stream with different lengths, i.e., document-summary pair number, as shown Figure~\ref{figures}(b).
The result shows that the ROUGE score of SSG increases with the number of input documents to begin with, which demonstrates that our proposed model is more and more effective when the text stream begins to grow larger.
It can also be seen that, as the stream gets longer, the improvement of our model compared with \texttt{BART-Split} is generally consistent.
Regardless of the different lengths of the text stream, our \texttt{SSG} consistently outperforms \texttt{BART-Split}, which is a strong baseline that achieves good performance in all metrics.
The above result demonstrates that the previous summary is always useful, and our model can successfully utilize it.

\subsection{Efficiency and Scale Analysis}

\textcolor{black}{From parameter view, both PEGASUS (223M) and our model (149M) introduce more parameters compared to the baseline BART, leading to increased computational resources. 
Secondly, a larger model introduces more training time and inference time. 
Hence, it takes a bit longer for our model to decode compared with BART (5s). 
Finally, the integration of the GAN mechanism in SSG, despite adding a modest increase in parameters (3.11M), significantly accelerates network convergence, and does not introduce more inference time.}

\textcolor{black}{Based on the findings in Tables~\ref{tab:together} and \ref{tab:comp_human_baslines}, it is evident that an increase in the number of parameters leads to improved model performance. 
For instance, GPT-3.5, with its larger parameter scale, achieves a human evaluation rating of 4.03.
At the same time, it is important to note that within the same parameter range, the design of the model significantly affects its performance. As an example, our model, which has only 6\% more parameters compared to BART-base, surpasses BART's performance by 12\%. 
Similarly, it surpasses PEGASUS by 4\% while having 40\% fewer parameters.}

\section{Conclusion}

In this paper, we have proposed a model named \emph{Stepwise Summary Generator} (SSG) which aims to generate summaries for stream document data.
Since previous information is also important when summarizing a current document, our model used a previous summary to selectively read the current document and obtained its polished representation, both were combined to generate the summary.
% Then, we propose to incorporate the polished representation of the current document and previous summary into the Transformer-based summary generator.
In addition, a convolutional discriminator was employed to provide a coherence signal to the generator, so as to enhance the coherence between the previous summary and the newly generated one.
SSG outperformed the state-of-the-art summarization models on our large-scale stepwise summarization dataset in terms of both human and automatic metrics.
In the near future, we will employ large language models to enhance the performance of the stepwise summarization models.

\section*{Acknowledgments}

We would like to thank the anonymous reviewers for their constructive comments. 
The work was supported by King Abdullah University of Science and Technology (KAUST) through grant awards FCC/1/1976-44-01, FCC/1/1976-45-01, REI/1/5234-01-01, RGC/3/4816-01-01, REI/1/5414-01-01, REI/1/5289-01-01, and REI/1/5404-01-01.

\bibliographystyle{unsrt}
\bibliography{reference}

\end{document}